\documentclass[runningheads]{llncs}
\usepackage{graphicx}
\usepackage{booktabs}
\usepackage{multirow}
\begin{document}
\title{Convolutional Neural Networks and a Transfer Learning Strategy to Classify Parkinson's Disease from Speech in Three Different Languages}
\titlerunning{CNNs and a Transfer Learning Strategy to Classify PD from Speech}
%
\author{J. C. V\'asquez-Correa$^{1,2}$, T. Arias-Vergara$^{1,2,3}$, C. D. Rios-Urrego$^{2}$, M. Schuster$^{3}$, J. Rusz$^{4}$, J. R. Orozco-Arroyave$^{1,2}$, E. N\"oth$^{1}$}
\authorrunning{J. C. V\'asquez-Correa et al.}
%
\institute{
	$^1$Pattern Recognition Lab. Friedrich-Alexander Universit\"at, Erlangen-N{\"u}rnberg, Germany\\
	$^2$ Faculty of Engineering. Universidad de Antioquia UdeA, Calle 70 No. 52-21, Medell\'in, Colombia \\
	$^3$ Department of Otorhinolaryngology, Head and Neck Surgery. Ludwig-Maximilians\\ Universit\"at, Munich, Germany\\
	$^{4}$ Department of Circuit Theory, Faculty of Electrical Engineering, Czech Technical University in Prague, Prague, Czech Republic\\
\email{juan.vasquez@fau.de}}
\maketitle              
\begin{abstract}

Parkinson's disease patients develop different speech impairments that affect their communication capabilities. The automatic assessment of the speech of the patients allows the development of computer aided tools to support the diagnosis and the evaluation of the disease severity. 
This paper introduces a methodology to classify Parkinson's disease from speech in three different languages: Spanish, German, and Czech. The proposed approach considers convolutional neural networks trained with time frequency representations and a transfer learning strategy among the three languages. 
The transfer learning scheme aims to improve the accuracy of the models when the weights of the neural network are initialized with utterances from a different language than the used for the test set. The results suggest that the proposed strategy improves the accuracy of the models in up to 8\% when the base model used to initialize the weights of the classifier is robust enough. In addition, the results obtained after the transfer learning are in most cases more balanced in terms of specificity-sensitivity than those trained without the transfer learning strategy.

\end{abstract}
\section{Introduction}

Parkinson's disease (PD) is a neurodegenerative disorder characterized by the progressive loss of dopaminergic neurons in the mid-brain producing several motor and non-motor impairments in the patients~\cite{hornykiewicz1998biochemical}. Motor symptoms include among others, bradykinesia, rigidity, resting tremor, micrographia, and different speech impairments. The speech impairments observed in PD patients are typically grouped as hypokinetic dysarthria, and include symptoms such as vocal folds rigidity, bradykinesia, and reduced control of muscles and limbs involved in the speech production.
The effects of dysarthria in the speech of PD patients include increased acoustic noise, reduced intensity, harsh and breathy voice quality, increased voice nasality, monopitch, monoludness, speech rate disturbances, imprecise articulation of consonants~\cite{tykalova2017distinct}, and involuntary introduction of pauses~\cite{Moretti2003}. Clinical observations in the speech of patients can be objectively and automatically measured by using computer aided methods supported in signal processing and pattern recognition with the aim to address two main aspects: (1) to support the diagnosis of the disease by classifying healthy control (HC) subjects and patients, and (2) to predict the level of degradation of the speech of the patients according to a specific clinical scale.

Most of the studies in the literature to classify PD from speech are based on computing hand-crafted features and using classifiers such as support vector machines (SVMs) or K-nearest neighbors (KNN). For instance, in~\cite{Sakar2013}, the authors computed features related to perturbations of the fundamental frequency and amplitude of the speech signal to classify utterances from 20 PD patients and 20 HC subjects, Turkish speakers. Classifiers based on KNN and SVMs were considered, and accuracies of up to 75\% were reported.
Later, in~\cite{Villa2015} the authors proposed a phonation analysis based on several time frequency representations to assess tremor in the speech of PD patients. The extracted features were based on energy and entropy computed from time frequency representations. Several classifiers were used, including Gaussian mixture models (GMMs) and SVMs. Accuracies of up to 77\% were reported in utterances of the PC-GITA database~\cite{OROZCOARROYAVE14.7}, formed with utterances from 50 PD patients and 50 HC subjects, Colombian Spanish native speakers.
The authors from~\cite{novotny2014} computed features to model different articulation deficits in PD such as vowel quality, coordination of laryngeal and supra-laryngeal activity, precision of consonant articulation, tongue movement, occlusion weakening, and speech timing. The authors studied the rapid repetition of the syllables /pa-ta-ka/ pronounced by 24 Czech native speakers, and reported an accuracy of 88\% discriminating between PD patients and HC speakers, using an SVM classifier.
Additional articulation features were proposed in~\cite{rafa2016book}, where the authors modeled the difficulty of PD patients to start/stop the vocal fold vibration in continuous speech. The model was based on the energy content in the transitions between unvoiced and voiced  segments. The authors classified PD patients and HC speakers with speech recordings in three different languages (Spanish, German, and Czech), and reported accuracies ranging from 80\% to 94\% depending on the language; however, the results were optimistic, since the hyper-parameters of the classifier were optimized based on the accuracy on the test set.  
Another articulation model was proposed in~\cite{moro2019forced}. The authors considered a forced alignment strategy to segment the different phonetic units in the speech utterances. The phonemes were segmented and grouped to train different GMMs. The classification was performed based on a threshold of the difference between the posterior probabilities from the models created for HC subjects and PD patients. The model was tested with Colombian Spanish utterances from the PC-GITA database~\cite{OROZCOARROYAVE14.7} and with the Czech data from~\cite{rusz2013imprecise}. The authors reported accuracies of up to 81\% for the Spanish data, and of up to 94\% for the Czech data.

In addition to the hand-crafted feature extraction models, there is a growing interest in the research community to consider deep learning models in the assessment of the speech of PD patients~\cite{grosz2015assessing,vasquez2017convolutional,Tu2017}. Deep learning methods have the potential to extract more abstract and robust features than those manually computed. These features could help to improve the accuracy of different models to classify pathological speech, such as PD~\cite{cummins2018speech}.
A deep learning based articulation model was proposed in~\cite{vasquez2017convolutional} to model the difficulties of the patients to stop/start the vibration of the vocal folds. Transitions between voiced and unvoiced segments were modeled with time-frequency representations and convolutional neural networks (CNNs). The authors considered speech recordings of PD patients and HC speakers in three languages: Spanish, German, and Czech, and reported accuracies ranging from 70\% to 89\%, depending on the language. However, in a language independent scenario, i.e., training the CNN with utterances from one language and testing with the remaining two, the results were not satisfactory (accuracy$<60\%$).

The classification of PD from speech in different languages has to be carefully conducted to avoid bias towards the linguistic content present in each language. For instance, Czech and German languages are richer than Spanish language in terms of consonant production, which may cause that it is easier to produce consonant sounds by Czech PD patients than by Spanish PD patients. Despite these language dependent issues, the results in the classification of PD in different languages could be improved using a transfer learning strategy among languages, i.e., to train a base model with utterances from one language, and then, to perform a fine-tuning of the weights with utterances from the target language~\cite{wang2015}. Similar approaches based on transfer learning have been recently considered to classify PD using handwriting~\cite{naseer2019}. In the present study, we propose a methodology to classify PD via a transfer learning strategy with the aim to improve the accuracy in different languages. CNNs trained with utterances from one language are used to initialize a model to classify speech utterances from PD patients in a different language. The models are evaluated with speech utterances in Spanish, German, and Czech languages. The results suggest that the use of a transfer learning strategy improved the accuracy of the models over 8\% with respect to those obtained when the model is trained only with utterance from the target language.

\section{Materials and methods}
\subsection{Data}
Speech recordings of patients in three different languages are considered: Spanish, German, and Czech. All of the recordings were captured in noise controlled conditions. The speech signals were down-sampled to 16\,kHz. The patients in the three datasets were evaluated by a neurologist expert according to the third section of the movement disorder society, unified Parkinson's disease rating scale (MDS-UPDRS-III)~\cite{goetz2008movement}. Table~\ref{tab:spkinf} summarizes the information about the patients and healthy speakers. 

\subsubsection{Spanish}
The Spanish data consider the PC-GITA corpus~\cite{OROZCOARROYAVE14.7}, which contains utterances from 50 PD patients and 50 HC, Colombian Spanish native speakers. The participants were asked to pronounce a total of 10 sentences, the rapid repetition of /pa-ta-ka/, /pe-ta-ka/, /pa-ka-ta/, /pa/, /ta/, and /ka/, one text with 36 words, and a monologue. All patients were in ON state at the time of the recording, i.e., under the effect of their daily medication.
\subsubsection{German}
Speech recordings of 88 PD patients and 88 HC speakers from Germany are considered~\cite{Skodda-2011}. The participants performed four speech task: the rapid repetition of /pa-ta-ka/, 5 sentences, one text with 81 words, and a monologue.
\subsubsection{Czech}
A total of 100 native Czech speakers (50 PD, 50 HC) were considered~\cite{rusz2018detecting}. The speech tasks performed by the participants include the rapid repetition of the syllables /pa-ta-ka/, a read text with 80 words, and a monologue.
\begin{table}[!ht]
	\centering
	\caption{Information of the speakers in the three datasets. \textbf{Subjects}: Number of speakers.
	\textbf{G.}: gender (\textbf{M.} male or \textbf{F.} female).
	\textbf{T}: Time after diagnosis in years. }
	\label{tab:spkinf}
	\resizebox{\linewidth}{!}{
\begin{tabular}{l|c|cc|cc|cc}
\toprule
             & \hspace{0.1cm} \multirow{2}{*}{\textbf{G}} \hspace{0.1cm} & \multicolumn{2}{c|}{\textbf{Spanish}}                                    & \multicolumn{2}{c|}{\textbf{German}}                                        &  \multicolumn{2}{c}{\textbf{Czech}}                                       \\
             & & \multicolumn{1}{c}{\textbf{PD}} & \textbf{HC}      & \multicolumn{1}{c}{\textbf{PD}} & \textbf{HC}        & \multicolumn{1}{c}{\textbf{PD}} & \textbf{HC}      \\
             \hline
\multirow{2}{*}{Subjects}     & M & 25                & 25                & 47                & 44                & 30                & 30                \\
             & F & 25                & 25                & 41                & 44                & 20                & 20                \\
             \hline
\multirow{2}{*}{Range of age} & M & 33-81             & 31-86             & 44-82             & 26-83             & 43-82             & 41-77             \\
             & F & 49-75             & 49-76             & 42-84             & 28-85             & 41-72             & 40-79             \\
             \hline
\multirow{2}{*}{Age}          & M & 61.3 (11.4)       & 60.5 (11.6)       & 66.7 (8.7)        & 63.8 (12.7)       & 65.3 (9.6)        & 60.3 (11.5)       \\
             & F & 60.7 \phantom{} (7.3)        & 61.4 \phantom{} (7.0)        & 66.2 (9.7)        & 62.6 (15.2)       & 60.1 (8.7)        & 63.5 (11.1)       \\
             \hline
\multirow{2}{*}{T}     & M & 8.7 \phantom{ } (5.9)         & --                 & 7.0 (5.5)         & --                 & 6.7 (4.5)         & --                 \\
             & F & 12.6 (11.6)       & --                 & 7.1 (6.2)         & --                 & 6.8 (5.2)         & --                 \\
             \hline
MDS-UPDRS     & M & 37.8 (22.1)       & --                 & 22.1 \phantom{} (9.9)        & --                 & 21.4 (11.5)       & --                 \\
-III             & F & 37.6 (14.1)       & --                 & 23.3 (12.0)       & --                 & 18.1 \phantom{} (9.7)        & --  \\
	\bottomrule
    \end{tabular}
	}
\end{table}

\subsection{Segmentation}
Speech signals are analyzed based on the automatic detection of onset and offset transitions, which model the difficulties of the patients to start/stop the movement of the vocal folds. The detection of the transitions is based on the presence of the fundamental frequency of speech in short-time frames, as it was shown in~\cite{rafa2016book}. The border between voiced and unvoiced frames is detected, and 80\,ms of the signal are taken to the left and to the right, forming segments with 160\,ms length. The transition segments are modeled with two different approaches: (1) a baseline model based on hand-crafted features, which are classified using an SVM, and (2) a model based on time-frequency representations used as input to train a CNN, which then will be used for the transfer learning strategy. Further details are given in the following subsections.

\subsection{Baseline model}
The features extracted from the transitions include 12 Mel-Frequency Cepstral Coefficients (MFCCs) with their first and second derivatives, and the log energy of the signal distributed into 22 Bark bands. The total number of descriptors corresponds to 58. Four statistical functionals (mean, standard deviation, skewness, and kurtosis) are  computed for each descriptor, obtaining a 232-dimensional feature-vector per utterance. The classification of PD patients and HC speakers is performed with a radial basis SVM with margin parameter $C=10$ and a Gaussian kernel with parameter $\gamma=0.0001$. The SVM is tested following a 10-fold Cross-Validation strategy, speaker independent.

\subsection{CNN model}
\label{sec:cnnmod}

Time frequency representations based on the short-time Fourier transform \linebreak (STFT) are used as input to a CNN, which extract the most suitable features to discriminate between PD patients and HC subjects. The STFT with 256 frequency bins is computed for each segmented transition, for a window length of 16\,ms and a step-size of 4\,ms, forming 41 time frames per transition. The obtained spectrogram is transformed into the Mel-scale using 80 filters, forming an spectrogram with a size of 80$\times$41, which is used to train the CNNs.
The architecture of the implemented CNN is summarized in Table~\ref{tab:cnnarch}. It consists of four convolutional and max-pooling layers, dropout to regularize the weights, and two fully connected layers followed by the output layer to make the final decision using a softmax activation function. The number of feature maps on each convolutional layer is twice the previous one in order to get more detailed representations of the input space in the deeper layers. The CNN is trained using the the cross-entropy as the loss function, using an Adam optimizer~\cite{adam}.

\begin{table}[!ht]
	\centering
	\caption{Architecture of the CNN implemented in this study.}
	\label{tab:cnnarch}
    \begin{tabular}{l|cc}
        \toprule
        \textbf{CNN Architecture} & \textbf{Input size} & \textbf{Output size} \\ \hline
        Conv (1x4x3,1)+dropout  &  1x80x41 & 4x80x41              \\
        Max Pool(2,2)           &  4x80x41 & 4x40x20              \\
        Conv (4x8x3,1)+dropout  &  4x40x20 & 8x40x20              \\
        Max Pool(2,2)           &  8x40x20 & 8x20x10              \\
        Conv (8x16x3,1)+dropout &  8x20x10 & 16x20x10             \\
        Max Pool(2,2)           &  16x20x10 & 16x10x5              \\
        Conv (16x32x3,1)+dropout &  16x10x5 & 32x10x5              \\
        Max Pool(2,2)            & 32x10x5 & 32x5x2               \\
        Lineal(320,128)+dropout  & 32x5x2 & 1x128              \\
        Lineal(128,64)+dropout   & 1x128 & 1x64               \\
        Lineal(64,2)             & 1x64 & 1x2                \\ \bottomrule
    \end{tabular}
\end{table}

\subsection{Transfer learning}
\begin{figure}[!ht]
\centering
	\includegraphics[width=\linewidth]{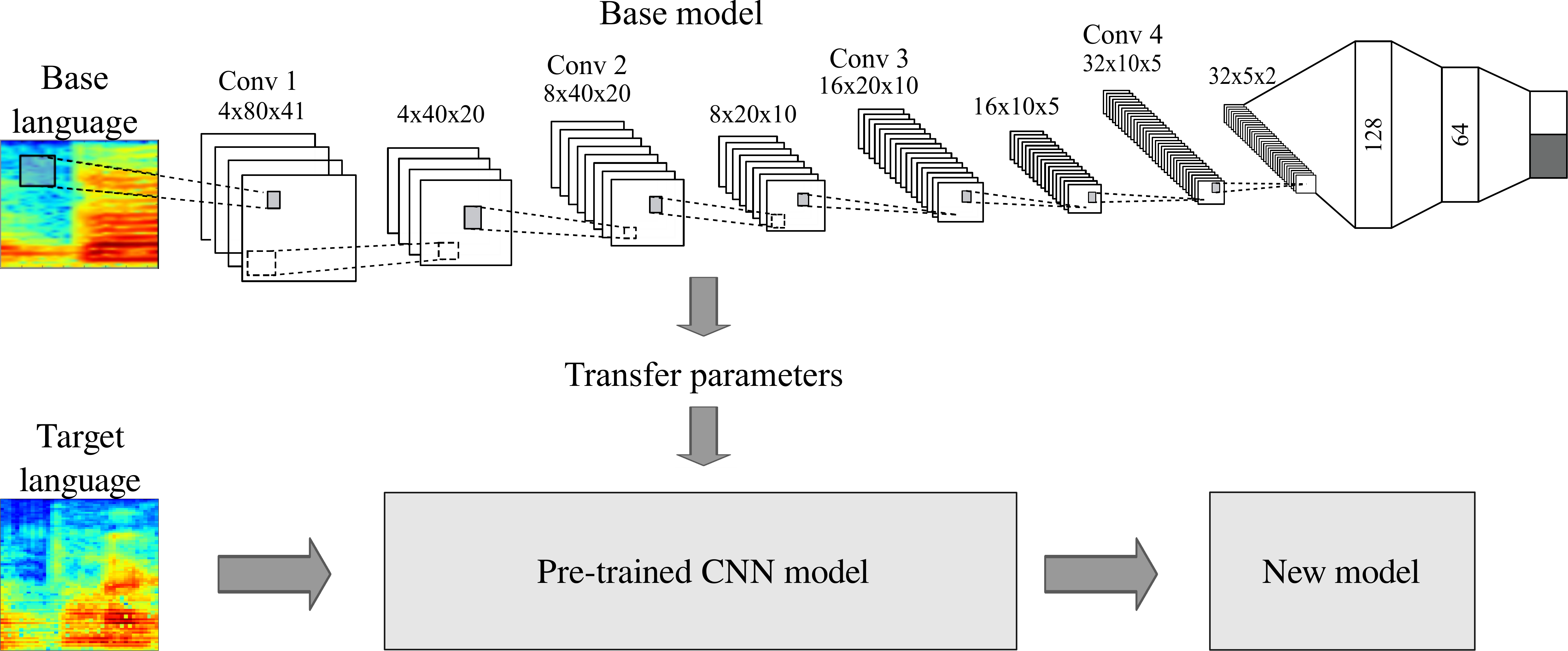}
	\caption{Transfer learning strategy proposed in this study to classify PD from speech with utterances from different languages}
	\label{fig:trlang}
\end{figure}

Transfer learning allows to use a neural network trained for one task to be used in another domain. We use transfer learning to classify patients and healthy speakers in three different languages. The CNN architecture described before is used to train a CNN with utterances from one language. Then, the pre-trained model is used as a base to initialize two different models with the remaining languages. Figure~\ref{fig:trlang} summarizes this procedure.

\section{Experiments and results}
The experiments are divided as follows: First, the baseline and the CNN models are trained considering each language individually. Then, the trained CNNs for each language are used as a base model in the transfer learning strategy in order to improve the accuracy in the other two languages. All speech exercises performed by the participants were considered for the classification strategy. The final decision for each speaker was obtained by a majority voting strategy among the different speech exercises. 

\subsection{Baseline and individual CNN models}
Table~\ref{tab:exp1} shows the results obtained for the baseline and the CNNs trained for each language individually. Similar accuracies are obtained  between the baseline and the CNN model for Spanish language, which also exhibit the highest accuracy among the three languages.  Note that the highest accuracy for German language was obtained with the baseline model. Conversely, for Czech language the CNN produces the highest accuracy. Note also that for the three languages, the results are unbalanced towards one of the two classes according to the specificity and sensitivity values.
The difference in the results obtained among the three languages can be explained considering the information provided in Table~\ref{tab:spkinf}. For the patients in the Spanish language, the average MDS-UPDRS-III score is higher compared with the German and Czech patients, i.e, there are patients with higher disease severity in the Spanish data compared to German and Czech patients.

\begin{table}[!ht]
	\centering
	\caption{Classification results for the baseline and CNN models trained in three different languages. \textbf{Acc}: Accuracy. \textbf{Sen}: Sensitivity. \textbf{Spe}: Specificity. \textbf{MCC}: Matthews correlation coefficient.}
	\label{tab:exp1}
	\resizebox{\linewidth}{!}{
    \begin{tabular}{l|cccc|cccc}
	\toprule
             & \multicolumn{4}{c}{\textbf{Baseline}}  & \multicolumn{4}{|c}{\textbf{CNN}}      \\
    \textbf{Language} &\textbf{ Acc (\%)} & \textbf{Sen (\%)} & \textbf{Spe (\%)} & \textbf{MCC} & \textbf{Acc (\%)} & \textbf{Sen (\%)} & \textbf{Spe (\%)} & \textbf{MCC}\\
    \hline
    Spanish  & 73.7 (13.0)      & 74.5 (16.7)            & 77.1 (16.2)   &0.50   & 71.0 (15.9)      & 74.0 (25.0)       & 68.0 (28.6)   & 0.42   \\
    German   & 69.3 \phantom{ }(9.9)       & 71.8 (12.4)            & 68.7 (10.0)    &0.39    & 63.1 (11.7)       & 43.1 (38.0)       & 83.1 (17.7)   & 0.30    \\
    Czech    & 61.0 (12.5)       & 64.5 (19.5)            & 60.2 (11.9)  & 0.27     & 68.5 (14.1)       & 94.0 (13.5)       & 42.0 (33.2) &  0.43 \\\bottomrule  
    \end{tabular}
    }
\end{table}
\subsection{Transfer language among languages}

The results with the transfer learning strategy among languages are shown in Table~\ref{tab:exp2}.  A CNN trained with utterances from the base language is fine-tuned with utterances from the target language.
Note that the accuracy improved considerably when the target languages are German and Czech, with respect to the results observed for baseline and the CNN in Table~\ref{tab:exp1}. The accuracy improved over 8\% for German (from 69.3\% in the baseline to 77.3\% when the model is fine-tuned from Spanish), and over 4.1\% for Czech language (from 68.5\% with the initial CNN to 72.6\% when the model is fine-tuned from Spanish). 
Particularly, the highest accuracy for German and Czech languages is obtained when the base language is Spanish. This can be explained considering that Spanish speakers have the best initial separability, thus, the other two languages benefit from the best initial model.  
The results obtained with the transfer learning strategy among languages are also more balanced in terms of the specificity and sensitivity than the observed in the baseline and with the initial CNNs. The standard deviation of the transfered CNNs is also lower, which leads to an improvement in the generalization of the models.

\begin{table}[!ht]
	\centering
	\caption{Classification results for the transfer learning among languages using CNNs. \textbf{Acc}: Accuracy. \textbf{Sen}: Sensitivity. \textbf{Spe}: Specificity. \textbf{MCC}: Matthews correlation coefficient. \textbf{Base lang}: Language used to pre-train the CNN model. \textbf{Target lang.}: Language used for transfer learning from the base model.}
	\label{tab:exp2}
    \begin{tabular}{l|c|cccc}
    \toprule
    \textbf{Base lang.} & \textbf{Target lang.} & \hspace{0.4cm} \textbf{Acc ($\%$)} \hspace{0.4cm} & \hspace{0.4cm} \textbf{Sen ($\%$)} \hspace{0.4cm} & \hspace{0.4cm} \textbf{Spe ($\%$)} \hspace{0.4cm}  & \hspace{0.2cm} \textbf{MCC}\\
    \hline
    German & \multirow{2}{*}{Spanish}      & 70.0 (12.5)       & 62.0 (19.9)        & 78.0 (23.9)  & 0.41     \\
    Czech &       & 72.0 (13.1)        & 67.0 (11.6)        & 78.0 (23.9)   & 0.46    \\
    \hline
    Spanish &  \multirow{2}{*}{German}      & 77.3 (11.3)       & 86.2 (13.8)       & 68.3 (14.3)  & 0.57     \\
    Czech &         & 76.7 \phantom{} (7.9)        & 87.5 (11.0)       & 66.0 (15.6)  & 0.55     \\
    \hline
    Spanish & \multirow{2}{*}{Czech}      & 72.6 (13.9)        & 82.0 (14.8)        & 62.0 (28.9)   & 0.46     \\
    German &          & 70.7 (14.5)       & 80.0 (16.3)       & 62.5 (26.3)    & 0.38    \\
    \bottomrule
    \end{tabular}
\end{table}

\begin{figure}[!ht]
    \centering
        \setlength{\tabcolsep}{0pt} 
\renewcommand{\arraystretch}{0} 
    \begin{tabular}{ccc}
       \hspace{0.3cm} \textbf{A.} & \hspace{0.3cm} \textbf{B.} & \hspace{0.3cm} \textbf{C.}\\
        \includegraphics[width=0.33\linewidth]{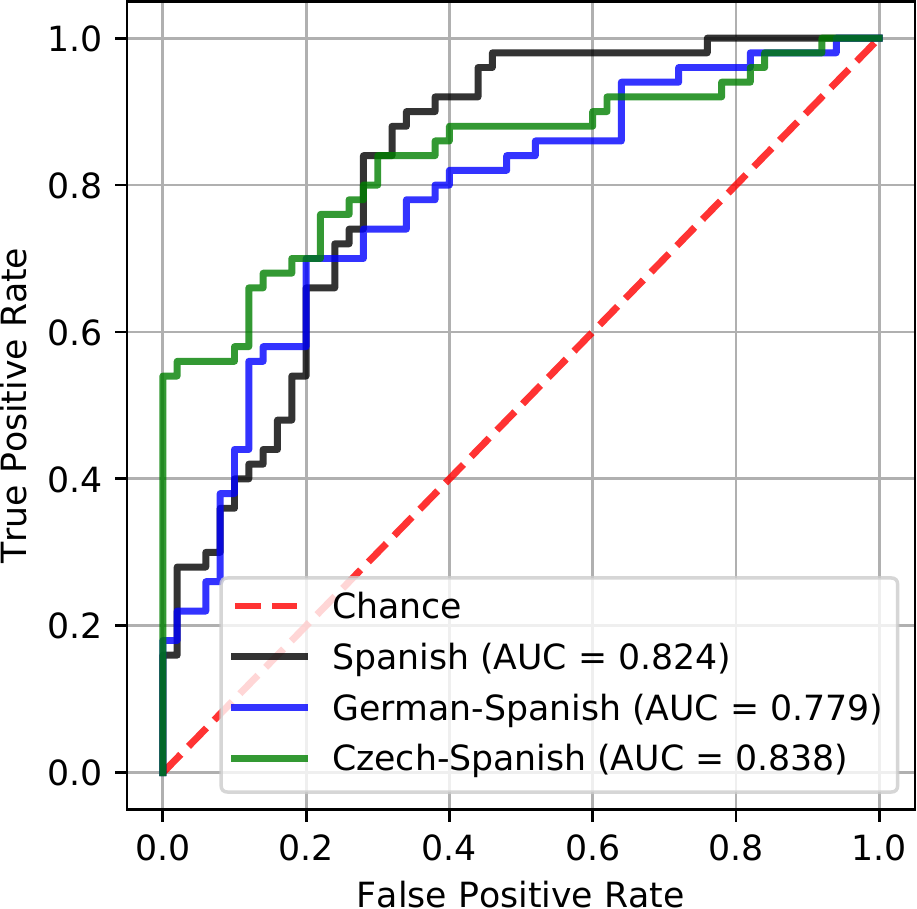} &         \includegraphics[width=0.33\linewidth]{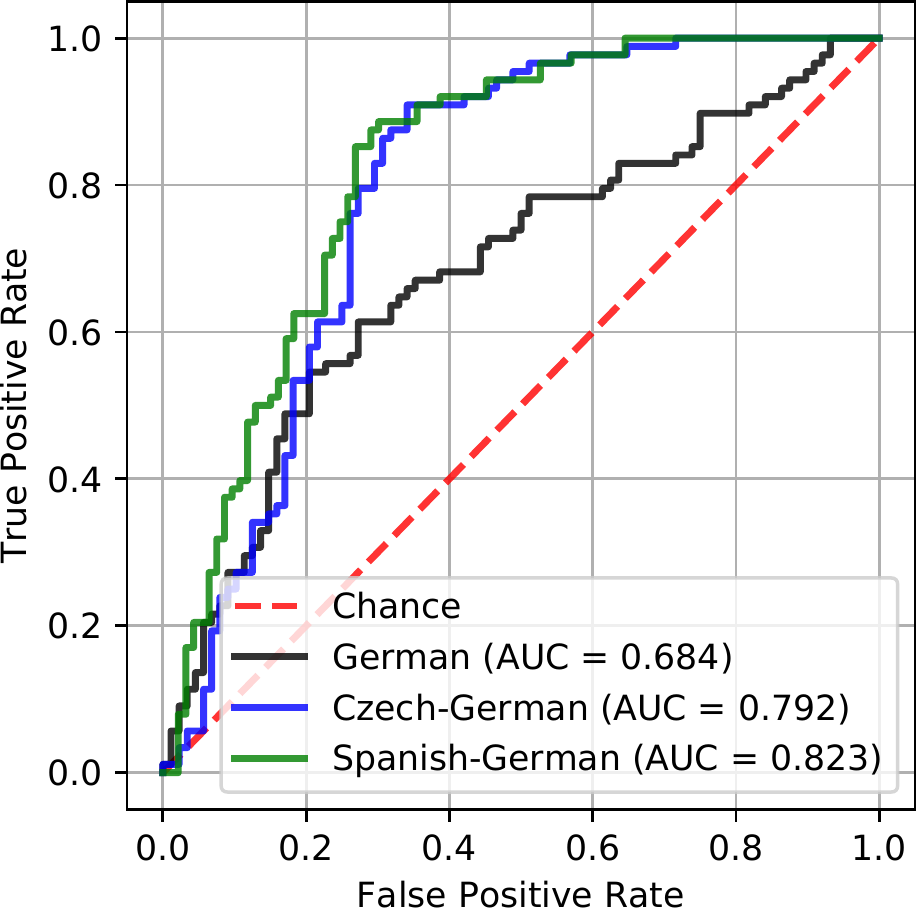} &         \includegraphics[width=0.33\linewidth]{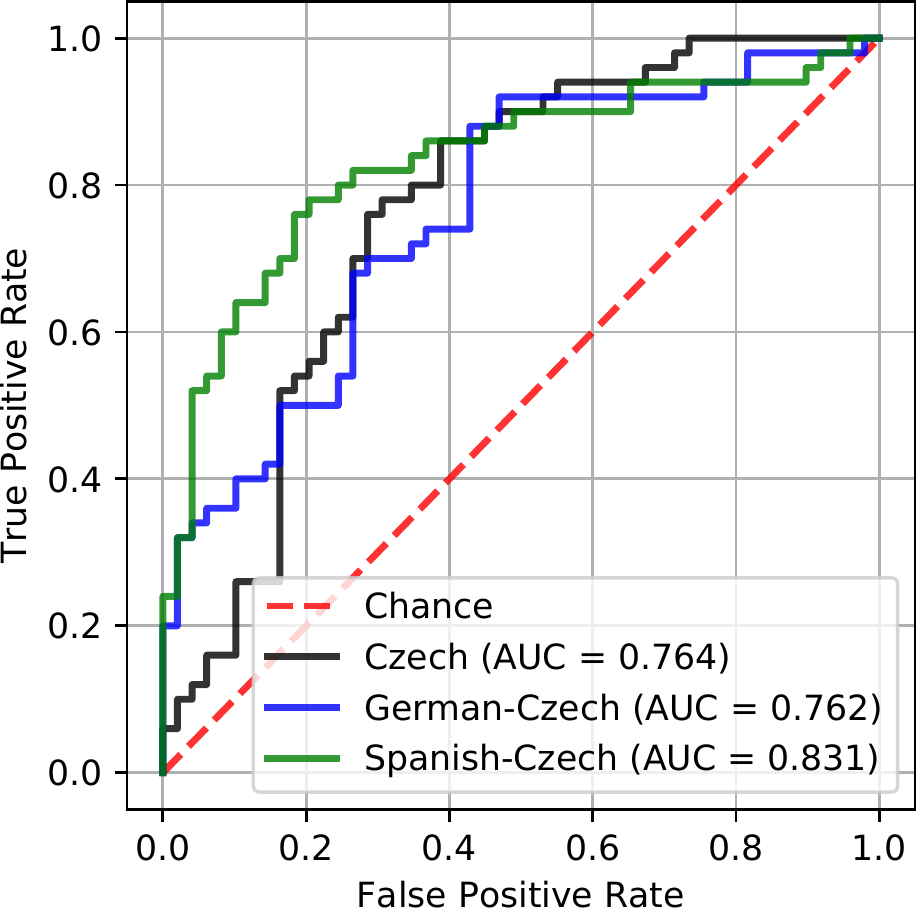}
    \end{tabular}
    \caption{ROC curves for the transfer learning among languages, when the target language is \textbf{A.} Spanish, \textbf{B.} German, and \textbf{C.} Czech.}
    \label{fig:rocs}
\end{figure}

The receiver operating characteristic (ROC) curves from Figure~\ref{fig:rocs} show with more detail the effect of the transfer learning strategy in the performance of the CNNs to classify PD speakers in different languages. The area under the ROC curve (AUC) when the target language is Spanish (Figure~\ref{fig:rocs}A) is slightly higher when the base language is Czech. When the target languages are German and Czech (Figure~\ref{fig:rocs}B and Figure~\ref{fig:rocs}C) the highest AUC is obtained when the base model is trained with Spanish utterances.

\section{Conclusion}

This study proposed the use of a transfer learning strategy based on fine-tuning to classify PD from speech in three different languages: Spanish, German, and Czech. The transfer learning among languages aimed to improve the accuracy  when the models are initialized with utterances from a different language than the one used for the test set.
Mel-scale spectrograms extracted from the transitions between voiced and unvoiced segments are used to train a CNN for each language. Then, the trained models are used to fine-tune a model to classify utterances in the remaining two languages.  

The results indicate that the transfer learning among languages improved the accuracy of the models in up to 8\% when a base model trained with Spanish utterances is used to fine-tune a model to classify PD German utterances. The results obtained after the transfer learning are also more balanced in terms of specificity-sensitivity and have a lower variance. In addition, the transfer learning among languages scheme was accurate to improve the accuracy in the target language only when the base model was robust enough. This was observed when the model trained with Spanish utterances was used to initialize the models for German and Czech languages. 

Further experiments will include the development of more robust base models using hyper-parameter optimization strategies like those based on Bayesian optimization. In addition, the base models will be trained considering two of the languages instead of only one of them. The trained models will also be evaluated to classify the speech of PD patients in several stages of the disease based on the MDS-UPDRS-III score, or based on their dysarthria severity~\cite{vasquez2018towards}. Further experiments will also include transfer learning among diseases, for instance training a base model with utterances to classify PD, and use such a model to initialize another one to classify other neurological diseases such as Hungtinton's disease.

\section*{Acknowledgments}
The work reported here was financed by CODI from University of Antioquia by grant Numbers  2017--15530 and PRG2018--23541. This project has received funding from the European Union’s Horizon 2020 research and innovation programme under the Marie Sklodowska-Curie Grant Agreement No. 766287. T. Arias-Vergara is also under grants of Convocatoria Doctorado Nacional-785 financed by COLCIENCIAS.
\bibliographystyle{unsrt}
\bibliography{references}
\end{document}